\documentclass{article}

\usepackage{arxiv}

\usepackage[utf8]{inputenc} % allow utf-8 input
\usepackage[T1]{fontenc}    % use 8-bit T1 fonts
\usepackage{hyperref}       % hyperlinks
\usepackage{url}            % simple URL typesetting
\usepackage{booktabs}       % professional-quality tables
\usepackage{amsfonts}       % blackboard math symbols
\usepackage{nicefrac}       % compact symbols for 1/2, etc.
\usepackage{microtype}      % microtypography
\usepackage{lipsum}		% Can be removed after putting your text content
\usepackage{graphicx}
\usepackage{natbib}
\usepackage{doi}

\title{\Large The Double-Edged Sword of Big Data and Information Technology for the Disadvantaged: A Cautionary Tale from Open Banking}

\author{ 
    {\hspace{1mm}Savina D. Kim} \thanks{S. Kim is affiliated with the University of Edinburgh Business School and the Centre for Technomoral Futures at the Edinburgh Futures Institute, United Kingdom} \\
	University of Edinburgh\\
	Edinburgh, United Kingdom \\
 	\And
    {\hspace{1mm}Galina Andreeva} \\
	University of Edinburgh\\
	Edinburgh, United Kingdom \\
	\And
    {\hspace{1mm}Michael Rovatsos} \\
	University of Edinburgh\\
	Edinburgh, United Kingdom \\
}

\date{}

\renewcommand{\shorttitle}{}

\hypersetup{
pdftitle={The Double-Edged Sword of Big Data and Information Technology for the Disadvantaged: A Cautionary Tale from Open Banking},
pdfsubject={stat.ML},
pdfauthor={Savina D.~Kim, Galina Andreeva, Michael Rovatsos},
pdfkeywords={Fairness, Discrimination, Financial vulnerability, Credit scoring, FinTech, Open Banking, Alternative credit data},
}

\begin{document}
\maketitle

\begin{abstract}
 This research article analyses and demonstrates the hidden implications for fairness of seemingly neutral data coupled with powerful technology, such as machine learning (ML), using Open Banking as an example. Open Banking has ignited a revolution in financial services, opening new opportunities for customer acquisition, management, retention, and risk assessment. However, the granularity of transaction data holds potential for harm where unnoticed proxies for sensitive and prohibited characteristics may lead to indirect discrimination. Against this backdrop, we investigate the dimensions of financial vulnerability (FV), a global concern resulting from COVID-19 and rising inflation. Specifically, we look to understand the behavioral elements leading up to FV and its impact on at-risk, disadvantaged groups through the lens of fair interpretation. Using a unique dataset from a UK FinTech lender, we demonstrate the power of fine-grained transaction data while simultaneously cautioning its safe usage. Three ML classifiers are compared in predicting the likelihood of FV, and groups exhibiting different magnitudes and forms of FV are identified via clustering to highlight the effects of feature combination. Our results indicate that engineered features of financial behavior can be predictive of omitted personal information, particularly sensitive or protected characteristics, shedding light on the hidden dangers of Open Banking data. We discuss the implications and conclude fairness via unawareness is ineffective in this new technological environment. 
\end{abstract}

% keywords can be removed
\keywords{Fairness\and Discrimination \and Financial vulnerability \and Credit scoring \and FinTech \and Open Banking \and Alternative credit data}

\section{Introduction}

Advances in information technology coupled with the availability of new data can fundamentally change the world for the better. Indeed, there are multiple examples of this already happening, from wearable devices and smart digital infrastructure helping improve outcomes of chronic disease patients \citep{Ghose2022} to online crowdfunding platforms providing small businesses with an alternative financing channel \citep{Luo2022}. Nevertheless, it is important to be aware of the unintended side-effects of any innovation, especially in relation to disadvantaged segments of the population. In this paper, we illustrate the potential risks of one such technology in the sector of household finance; specifically, we highlight the hidden associations of seemingly neutral Open Banking (OB) transactions with sensitive and protected personal characteristics of the financially vulnerable. To the best of our knowledge, this is the first study where solid evidence for this is given using real OB data.

With growing enthusiasm around machine learning (ML) and alternative credit data over the last few years, the implementation of the second Payment Services Directive (PSD2 or Open Banking in the UK\footnote{On October 27, 2022, the Consumer Financial Protection Bureau (CFPB) announced the introduction of Open Banking in the USA. This announcement extends a worldwide trend in opening financial transactions data which makes our investigation even more timely and relevant.}) is a game changer, particularly in credit risk assessment \citep{Remolina2019}. It enables easier accessibility to dynamic, real-time consumer data in practice, with bank transaction data being of particular interest. However, this rapid data shift and fast-paced development of algorithmic decision-making has simultaneously raised several social and ethical concerns. This is worsened by the fact that regulations "have not kept pace with modern Big Data capabilities” \citep[p. 24]{Wolkowitz2015} where the use of expanded data types is still porously regulated and largely unbounded. Questions arise regarding the magnitude of power of OB data in customer profiling, and whether one should be cautious from the standpoint of fairness and equality principles \citep{Hiller2022}. 

Fairness, and especially bias, in data and algorithmic decision-making has recently received a lot of attention from information science academics and practitioners \citep{Bellamy2019, Chouldechova2018, Lessmann2015, Mehrabi2019, corbettdavies2017}. This paper extends this line of inquiry by examining the associations of sensitive and legally protected personal characteristics with seemingly neutral financial transactions. In this study, we look to better understand the risks concealed in novel types of data which may go unnoticed. In doing so, we simultaneously provide a warning to all data modelers, and especially to lenders who are too hastily taking an “all data is credit data” \citep{Aitken2017} approach without proper caution. We investigate these risks in the context of financial vulnerability (FV), a common denominator across those in need of credit as well as being an urgent, global concern. Due to the unprecedented economic impact brought by the COVID-19 shock and further exacerbated by double-digit inflation and a looming energy crisis, a sharp rise to living costs has dramatically accelerated financial distress worldwide. In the UK alone, the number of adults with low financial resilience increased by 3.5 million between March and October of 2020, according to the Financial Lives survey \citep{fca2020survey} with similar conclusions made by Lowell’s Vulnerability Index which indicated a 11\% rise in financial vulnerability (FV) across the pandemic \citep{Braga2021}. A survey in the US revealed significant inequality in financial impact of COVID-19, where those most financially vulnerable prior to the pandemic faced even greater financial strain \citep{Bruce2022}. 

Surprisingly, there is no research connecting FV with the objective financial data, as previous limited research relied on self-reported FV measures from surveys. To fill this gap, we provide a comprehensive investigation of FV dimensions and their drivers using information derived from bank transaction data. In doing so, we demonstrate the power of OB data in profiling FV, which can be used for social good to help the disadvantaged segments. On the other hand, the same power can be abused e.g., by predatory lenders targeting customers who cannot afford credit repayments \citep{Jones2020}. 

The analysis is performed on a large OB dataset containing 180 million bank transaction records from up to 100,000 individuals in the UK sourced from a social FinTech lender. This unique dataset also contains some socio-demographic variables, including gender, which is not normally collected by lenders, since it is one of the characteristics not allowed for making credit-granting decisions.\footnote{The data provider collects gender for Know Your Customer (KYC) purposes, and does not use it for credit decisions.}  From financial transactions we can infer another protected characteristic – disability, and two additional ones that can be considered sensitive, namely whether the account-holder is a carer and if s/he has children. 

Drawing from regulatory guidance \citep{fca2015}, we propose six binary data-driven FV indicators (FVI) that can be used to identify vulnerable customers. We then generate a wide range of financial behaviors from financial management to financial inclusion that may be associated with FV as suggested by regulations and previous research. Knowledge of which behaviors may lead to FV can help in designing appropriate interventions. Bivariate Pearson linear correlation reveals statistically significant associations not only between FVI and financial behaviors, but also with protected characteristics. We use state-of-the-art machine learning algorithms to predict FVIs from financial behaviors and compare predictive accuracy as measures of successful performance. This approach has an advantage over Pearson’s correlation coefficient in capturing complex non-linear associations with multiple predictors. We also predict protected characteristics from financial behaviors and observe very high predictive accuracy, signaling powerful associations, and demonstrating the power of OB data. 

We then illustrate the potential biases that may arise through unawareness of these correlations. We show how segmentation and clustering (a popular tool in marketing and customer management) can implicitly capture sensitive and protected characteristics, even when they are not included in the data and its analysis. Furthermore, we reveal that clusters are characterized by the combination of several sensitive characteristics. Therefore, caution should be taken in making decisions not only with regards to a particular individual sensitive or protected characteristic, but also in terms of their combinations.

The contributions of this paper can be summarized on two levels. First, on a more general level, we contribute to the body of research on responsible and ethical AI through the lens of a real-world scenario and a practical application with lessons learned for fair modeling practices. The hidden connections with protected and sensitive characteristics can be present in any data across various applications, thus modelers and end users should be aware of this risk. We also extend the discourse on data biases by looking at the new types of alternative credit data that have not been analyzed previously, particularly granular transaction data. Second, at the context-specific level, we make several important contributions by quantifying the dimensions and drivers of FV. This is accomplished by demonstrating the predictive power of OB data in modeling FV and by proposing a data-driven segmentation of financial behaviors for enhanced applicant profiling.

The remainder of this paper is structured as follows. Section \ref{sec:Related Work} summarizes the relevant studies, including the impact of Open Banking, an examination of fairness concerns in credit systems and FV assessment methods. Section \ref{sec:Methods} introduces the FV indicators, followed by the empirical results of the ML-based prediction and clustering methods in Section \ref{sec:Results}. The final Section \ref{sec:Discussion} concludes the paper with a discussion of the findings, its implications for public policy makers and practitioners, and limitations which offer promising opportunities for future research. 

\section{Related Work}
\label{sec:Related Work}

\subsection{Open Banking}

Open Banking refers to a regime in which banks provide access to customer financial transactions in secure, digitalized form – with customers’ consent – to authorized third-party service providers such as FinTech companies. Standardized open application programming interfaces (APIs) are used by third parties to deliver services to customers using their own data, ranging from money management applications and financial product comparisons to applications for loans or mortgages. This has revolutionized financial services, pushing the boundaries of traditional credit risk assessment (i.e., credit scores), which often fail to identify more nuanced behaviors leading up to financial difficulty and rely on static, outdated data pulls rendering them opaque and error prone \citep{oleary2021}. This, coupled with the growing power of computational tools and reduction in data storage costs, has opened a new frontier for extracting inference from big data \citep{cortereal2017}.

In particular, transaction records can reflect a consumer’s risk profile on the basis of historical and current financial habits and preferences. \citet{vissing2012} argues there is informative value in what consumers buy when predicting default risk; for example, spending on entertainment such as magazines or toys. Customers’ personal information and aggregated transaction history on annual and monthly bases have also been shown to be effective \citep{Zhang2018} along with the balance of checking and savings accounts and cash inflows and outflows \citep{Khandani2010}. Even grocery shopping data has been shown to be informative when predicting credit card repayment behaviors \citep{Lee2021}. However, these studies focus on only creditworthiness, skipping over its major determinants such as financial health, stability and, correspondingly, vulnerability. Our study seeks to close this gap. To summarize, alternative data enables financial institutions to gain a more holistic view of an individual’s financial health and can help overcome the limitations of traditional methods; however, its intrusiveness simultaneously increases potential for individual harm. Therefore, how to guarantee consumer priorities such as fairness is still an open question, noting the minimal research on Open Banking data in relation to fairness considerations needed for its ethical use. These concerns are discussed in detail next.

\subsection{Discrimination in Credit and Algorithmic Bias}

One major concern around the use of alternative data for predicting individual levels of risk is enabling predatory lenders to identify vulnerable groups more easily and further perpetuating the harmful cycle of discrimination \citep{Hiller2022}. In other words, the same technology designed to make the “credit invisibles” of contemporary financial markets visible, is the same technology with the ability to precisely identify them \citep{Jones2020}. Prominent examples of this occurring are the targeting of minority borrowers for high-interest, subprime loans by fringe lenders making it harder for them to build a strong history of repayment; or the historic practice of redlining, resulting in a vicious cycle of restricted access which dramatically impacts communities of color \citep{Hiller2022}. According to sociological investigations by \citet{Fourcade2017}, segmentation methods increase social stratification and subjectification in terms of access to consumer credit in the American market, negatively impacting the life chances of select individuals. Similar conclusions were drawn by others when analyzing the technological disruption of the fringe finance infrastructure \citep{Langevin2019}. This risk is magnified by findings which show that offering financial incentives (e.g., lower interest rates) drives many to disclose personal information with minimal consideration of the possible consequences to such data sharing \citep{Acquisti2013, Norberg2007}, risking the transition of unsound fringe lenders into mainstream lenders. \citet{Packin2016} also explore potential consequences of emerging social credit systems, which authorize the use of highly personal information in return for better interest rates. They sound the alarm claiming direct and derivative harms to loan seekers regarding privacy, social segregation and due process violations derived from unsupervised ML. 

Facially neutral features may also be highly correlated with a protected characteristic (e.g., race or disability). A predictive model works by capturing all the features characterizing an event which is then utilized to make predictions; however, certain features may not only characterize the intended event but also inform another phenomena or class, also known as proxies \citep{Veale2017}. For example, factors such as educational background, wealth, work history, geography or even where one goes grocery shopping can be used to infer an individual’s race and thus exploited without proper regulation \citep{Hurley2017}. Furthermore, because fitting the majority population is more important for reducing overall modelling error, this leads to different (and often higher) distribution of errors for the minority population, who are thus systemically handicapped to begin with \citep{Chen2018, Chouldechova2018}. ML techniques are designed to fit the data therefore it is expected that they will replicate and amplify any bias already existing in the data; we have no reason to expect them to remove it. Therefore, utilizing unbalanced data can be particularly harmful towards underrepresented (and underfinanced) groups, which is often the case in financial services \citep{Rovatsos2019}.

Alongside embedded biases in credit scoring models, economic hardships are also reported at higher rates for racial and ethnic minorities, making them particularly a target. Black (73\%) and Hispanic adults (70\%) reported that they lacked emergency funds to cover three months of expenses (compared to 47\% White adults), and they more often reported (48\% and 44\%, respectively) that they would be unable to fully pay their bills during the recent COVID-19 pandemic, a real-world example of financial shock \citep{Lopez2020}.

\subsection{Financial Vulnerability}

FV, often used interchangeably with financial fragility, distress, debt burden, and overindebtedness, refers to an individual’s ability to manage daily finances, their resilience to economic shocks (e.g., unexpected rent increase or medical expense), and their capacity to pursue financial opportunities. Conceptually, it is a multi-faceted concept making it difficult to define and thus quantify \citep{perrig2016}. To help financial institutions and policymakers identify and support these ‘at-risk’ individuals, recent studies have looked to define formative measures of FV in hopes of attenuating its negative impact on financial outcomes. 

On an empirical basis, analysis of objective financial outcomes, or those straightforwardly derived from bank account records, are common. These include assessing one’s ability to pay their debts, such as debt-to-income ratio \citep{Costa2012} or timely repayment, when debt is in arrears for more than 90 days \citep{Il2016}. Other indicators measure income, expenses, and wealth levels \citep{Ampudia2016}, including whether cash flow can cover basic living costs such as utility bills \citep{Bridges2004}. Being unable to take a vacation, going out for a meal with friends, or enjoying leisure activities are also considered \citep{Worthington2006}. Subjective approaches using self-reported surveys are also used to capture individuals’ own perception of their financial situation. Whereas more difficult to scale and interpret, they capture a unique aspect of consumer welfare, such as one’s anxiety over personal finances, which purely objective financial data cannot. For example, \citet{Lusardi2011} ask individuals how confident they are that they could come up with \$2000 in 30 days to face an unexpected need. Policy-making bodies and regulators have also begun to pay greater attention to the subjective dimension, evidenced by the Financial Lives surveys \citep{fca2017, fca2020survey} and the measures for financial wellbeing \citep{cfpb2015}. Other research streams have examined the impact of behavioral patterns on FV likelihood, including money management skills, personal savings orientation, consideration of future consequences \citep{Rustichini2016} as well as psychological characteristics \citep{Gladstone2019}. It is argued that financial illiteracy and lack of self-control are major determinants \citep{Gathergood2012, Lusardi2019}, evidenced by high levels of overindebtedness in the most vulnerable individuals with multiple sources of (often unsecured) debt. It seems that behavioral biases, such as impulsivity, may lead to spending beyond one’s means as they are unable to resist instant gratification \citep{DeHart2016}. On the other hand, to meet necessary living costs under income shocks, individuals may also have no choice but to rely on resources outside of their savings, pensions or benefits, namely, consumer credit. However, vulnerable individuals often lack or have insufficient credit history resulting in financial exclusion and deprivation, particularly from traditional financial institutions \citep{Brevoort2016}. We rely on these findings and insights when defining features, indicators, and determinants of FV as described in the next section. 

\newpage
\section{Data and Methods}
\label{sec:Methods}
\subsection{Data Description}

\begin{figure}[h]
\includegraphics[width=\linewidth]{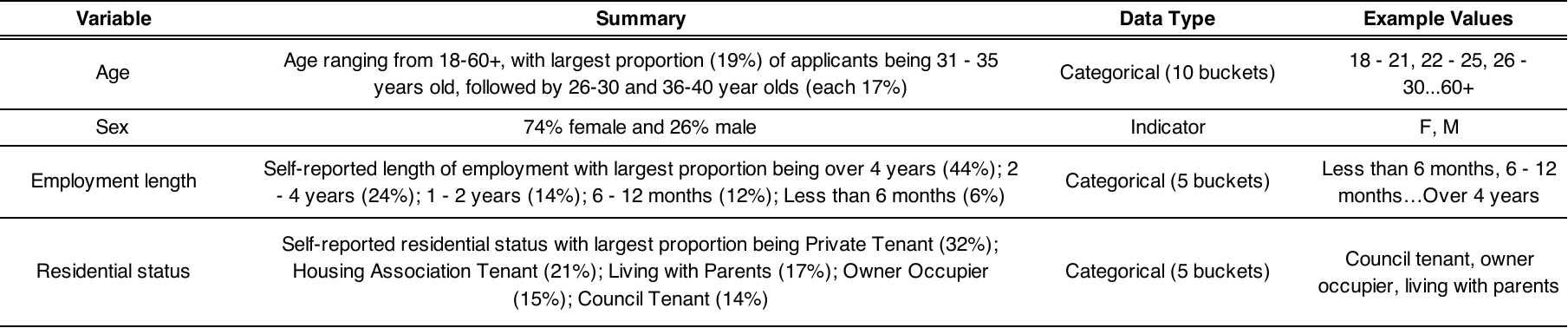}
\centering
\caption{Details of variables used from applicants’ demographic information data.}
\label{fig:fig1}
\end{figure}

We leverage a new, proprietary dataset provided by a UK-based social lender that primarily lends to individuals working in the public sector with loan sizes typically ranging between £500 to £1,000. The lender is unique in that they assess applicants via Open Banking during the affordability check procedure, excluding the need for a credit score. Using a third-party Open Banking API, the lender aggregates transactions across the applicant’s current account. The data was collected in February 2022 and focuses on approximately 100,000 applicants who have applied for a loan in the previous two years, yielding a dataset of over 180 million transactions. All applicant information is anonymized and includes demographic variables as illustrated in Figure \ref{fig:fig1}.

Historical bank transaction records of each applicant include transaction amount in British pounds (GBP), date, description, classification, category and remaining account balance. The data provider uses a third-party algorithm to categorize transactions into 56 categories (e.g., groceries \& housekeeping, earnings, etc.) with transaction references providing additional color on the merchant or service provided. To ensure a sufficient level of data per applicant and capture those actively using their account, we retain applicants who exhibit complete transaction details for a minimum observation period of six months and with at least ten transactions per month. 

\subsection{Financial Vulnerability Indicators}

There is no official or universally accepted definition of FV. Since the data is from the UK context, we refer to the Financial Conduct Authority’s (FCA) guidance on financial difficulties \citep{fcaconc1.3}. We consider key indicators of being ‘financially in difficulty’ or having ‘low financial resilience’ which include having insufficient funds in their account, being over-indebted, having low or erratic incomes or low savings and being unable to withstand an unexpected increase in monthly expenses such as rent. In line with the FCA, six target binary variables/key FV indicators are derived to measure applicants’ status and include the following. 

\begin{itemize}
	\item \emph{(1) Financial shock withstanding} (48.3\% of applicants): When an applicant is unable to withstand the impact of an unexpected expenditure of £100 on their account for more than 50\% of the months throughout their account history as included in the training or test samples. This is computed by referencing the average monthly median of an applicant’s account balance, where a median value less than £100 would suggest the applicant would struggle to sustain a shock. Since we only have access to transaction data from current/checking accounts, we are unable to view savings amounts, but this is likely to be minimal given only 4\% of applicants show evidence of payments into the category of \emph{savings and investments}. Additional threshold levels, including when individuals can withstand financial shock 0\% of all months and thus extremely financially vulnerable (\emph{“financial shock withstanding never”} comprising 18.4\% of applicants) and when individuals can withstand financial shock 100\% of all months and thus financially healthy (\emph{“financial shock withstanding always”} comprising 19.0\% of applicants), are assessed to represent more extreme cases. This enables us to compare across different intensities of FV and determine if select behaviors are monotonic.
	\item \emph{(2) Insolvent} (4.0\% of applicants): When an applicant has made at least one or more payments under the transaction category of debt management and insolvency in their history. This indicates an applicant has not been able to pay their debts when they were due and therefore overindebted.
	\item \emph{(3) Insufficient disposable income} (11.6\% of applicants): When an applicant has less than or equal to £100 in average monthly disposable income. To determine whether individuals can cope with a financial shock, we assess whether they can afford their necessary expenditures or would be deprived.
 \item \emph{(4) Overdraft} (67.4\% of applicants): When an applicant has at least one or more days in overdraft (OD) per month for more than 50\% of the months throughout their account history. OD indicates insufficient funds in the account. Other thresholds where individuals are in OD 0\% of months and thus financially healthy (\emph{“overdraft never”} comprising 6.9\% of applicants) and individuals are in OD 100\% of months and thus extremely financially vulnerable (\emph{“overdraft always”} comprising 35.1\% of applicants) are also assessed.
 \item \emph{(5) Returned direct debits} (28.3\% of applicants): When an applicant has at least one or more returned direct debits (RDD) on average per month. This value indicates insufficient funds resulting in a rejection of a pre-arranged payment by the bank.
 \item Another feature of interest, which does not directly imply FV, however, may be related to managing financial matters, includes being a \emph{(6) Gambler} (20.2\% of applicants). An applicant is considered a gambler if they have spent £100 or more on average per month on gambling expenditures. We propose that it is the compilation of indicators which constitute the umbrella term FV.
\end{itemize}

\subsection{Protected and Sensitive Attributes}

Central to this study are legally protected characteristics given their importance in anti-discrimination laws, also referred to as fair lending laws, such as the Fair Credit Reporting Act (FCRA), Equal Credit Opportunity Act (ECOA) and the Fair Housing Act (FHA) in the United States, the Equality Act in the United Kingdom and similar consumer protection directives in the European Union with Articles 12 and 13 of the EC Treaty which prohibit and provide measures to combat discrimination, respectfully. These make up the most significant legal instruments governing consumer credit scoring regulation which outline processes by which fairness should be met, most notably by prohibiting discrimination by racial or ethnic origin, nationality, religion or belief, sex, age, disability, or sexual orientation, and being used as basis for decision-making. 

We also consider some additional characteristics that may be seen as sensitive, especially if used in decision-making. These characteristics are derived from the FCA guidance, which describes a list of risk factors common to vulnerable consumers who, due to personal circumstances, are more susceptible to financial detriment \citep{fca2015}. These risk factors are also highlighted in consumer and public policy research \citep{Anderson2018, Griffiths2011, Moschis2011} and include: low education and financial literacy, physical disabilities, severe or long-term illnesses, mental health issues, low income, high debt, caring responsibilities, being either “young” or “old,” lack of English language skills, and impactful changes in personal circumstances, such as a divorce, death of a spouse, or a redundancy. Not every individual falling into one or more of these categories will necessarily experience FV, however, these factors are expected to increase the susceptibility of entering a financially vulnerable state as well as experiencing the severity of its consequences. 

Taking both risk factors and protected characteristics into consideration, our work examines select attributes that are available in our data. We refer to these as ‘socio-demographic profile’ features which describe an applicant and are commonly considered influential factors in their behavior and financial status. The lending company obtains socio-demographic characteristics at origination, including gender and age, alongside what we infer as sensitive attributes. These include whether an applicant has a disability (8\% of applicants), which is determined using disability benefit payments as a proxy, and attributes such as having a child (37.5\% of applicants) or being a carer (2.5\% of applicants), which are detected on the basis of receiving child and carer benefit payments, respectively. 

\subsection{Feature Engineering}

We identify a spectrum of data elements characterizing an applicant’s financial behavior next. Feature engineering is used to construct metrics which define and measure those behaviors to create a ‘financial profile.’ The transactions enable the construction of monthly inflows and outflows such as salary and benefits received, consumption patterns in relation to financial management ability (e.g., gambling habits, loan repayments, etc.) as well as temporal features such as volatility. We construct features across six proposed categories: (1) financial management, (2) financial distress, (3) financial resilience, (4) financial planning, (5) financial aid and (6) financial inclusion. Taken together, these enable applicant characterization based on level of financial health, degree of financial management ability and overall stability. The formulation of each category is detailed in the following Figure \ref{fig:fig2}.

\begin{figure}
\includegraphics[width=14cm]{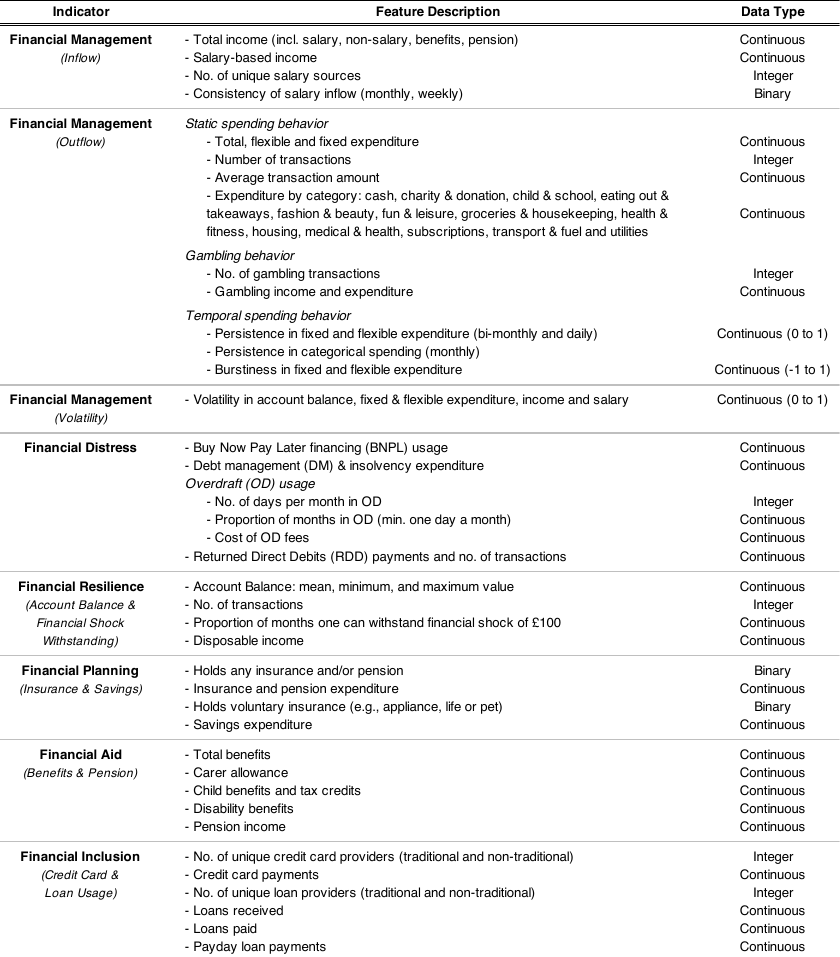}
\centering
\caption{Financial management and financial difficulty features engineered from transaction data. \emph{Note: All amount-based attributes are calculated as monthly averages across the applicant's account history in £/GBP unless noted otherwise.}}
\label{fig:fig2}
\end{figure}

\subsubsection{Financial Management}

We define summary statistics of applicants’ \emph{inflow} behavior as the average monthly total income, average monthly salary and non-salary income, number of unique salary sources and consistency of salary inflow (monthly or weekly) of each applicant. Income is computed as all major inflows into the applicants’ bank account, which includes, salary-based income, non-salary-based income, benefits received, pension received, and loans received, found under the categories: \emph{earnings} and \emph{credit bank transfers}. However, transactions considered as \emph{internal transfers}, \emph{returned direct debits} and \emph{returns} are excluded from income calculation. To further distinguish between salary-based and non-salary income, transaction references containing select non-salary patterns (e.g., mobile transfers and gambling-related inflows) are excluded as well as transaction amounts less than £100 and multiples of £10, as these are majorly small money and/or mobile transfers made from other individuals or family members. Salary and non-salary income are differentiated because in the presence of liquidity constraints, another important resource individuals may consider are loans or gifts from family and friends, which can ensure minimum levels of consumption \citep{midoes2022}. While these loans are typically short-term and small \citep{Long2020}, they do provide an additional financial buffer. Therefore, formal and informal credit options are both acknowledged to better assess FV. The number of unique salary sources is calculated using the references of salary-labeled transactions where unique references are counted as distinct sources. Salary inflows are labeled as consistent if more than 70\% of an applicant’s salary-based income is received in the same 10-day window every month (monthly) or three-day window every week (weekly).

\begin{figure}[h]
\includegraphics[width=9.5cm]{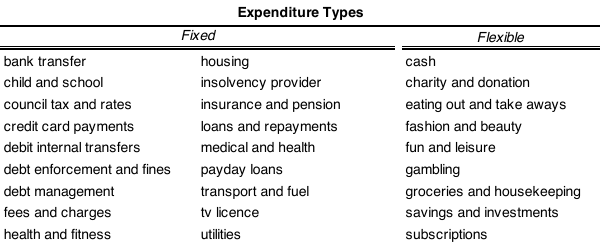}
\centering
\caption{List of categories organized as fixed versus flexible expenditures.}
\label{fig:fig3}
\end{figure}

We define summary statistics of applicants’ \emph{outflow} behavior as the average monthly total expenditure, including fixed and flexible expenditures (Figure \ref{fig:fig3}), number of transactions, average transaction amount, and monthly expenditure by spending category including \emph{cash, charity and donation, child and school, eating out and takeaways, fashion and beauty, fun and leisure, groceries and housekeeping, health and fitness, housing, medical and health, subscriptions, transport and fuel} and \emph{utilities}. Given the association between gambling, addiction and financially harmful outcomes, gambling expenditures are considered separately. This includes metrics for average number of gambling transactions per month and average monthly gambling income and expenditure. To identify gambling transactions, Natural Language Processing (NLP) techniques of the transaction reference are used; further details can be found in the Appendix (Figure A.1). 

Measures of persistence, burstiness and volatility are calculated next in order to consider temporal patterns. \emph{Persistence} is used to evaluate the consistency in the amount an applicant spends in a monthly and weekly observation period $t^\prime$ = \{M, W\}; this metric is computed using the average cosine similarity coefficients between adjacent time intervals. For the monthly observation period, first bi-monthly spending (i.e., two elements for each month) are aggregated followed by the fraction of spending in each element. The persistence in spending amount is then calculated as the average of the cosine similarity:

\begin{equation}
{persistence}_{Monthly}=\ \frac{\mathrm{\Sigma}_{i=0}^{n-1}cos{{(S}_i,S_{i+1})}}{n}
\end{equation}

where $S_i$ represents the vector of the relative amount spent in each bi-monthly (two-week) interval in a particular month i, and n represents the number of months we have of each applicant. This enables comparison between the first half of the month to the first half of the next month and subsequent months. A persistence value of zero implies that the relative amounts spent are dissimilar between the time intervals, whereas a value of one indicates they are the same across the time intervals. Similarly, weekly persistence is also calculated by grouping the spending amounts on a daily basis (i.e., 7 elements for each week). This enables comparison between consumption on a Monday one week and a Monday in subsequent weeks, for example. Persistence is calculated for both fixed and flexible expenditure amounts as well as the amount spent in each spending category; for example, to determine if an applicant consistently spends similar amounts in \emph{child and school} items every month. These features are derived to gain a sense of diversity and stability within each spending category over time. However, because the value of transactions can be biased towards high-value categories (i.e., purchasing a washing machine), metrics based on the total number of transactions, as opposed to total value of the transactions, are also calculated to measure the frequency of purchasing activities across various categories.

Bursty dynamics, or \emph{burstiness}, is defined as the heterogeneous property of time series which have short periods of intense high-frequency activities alternating with long periods of low-frequency activities \citep{Tovanich2021}. This metric is used to measure the intensity of expenditure patterns. Burstiness is computed by first taking inter-event times, or the daily difference between two subsequent transactions. In our case, the transaction date is considered given time of purchase is not available. The inter-event time is defined as $t_i$ = $T_i$ – $T_{i-1}$ where $T_i$ represents the transaction conducted at time i. The burstiness parameter is calculated as:

\begin{equation}
    B=\ \frac{r-1}{r+1}\ \ \ \ \ \ \ \ \ ,\ \ r=\ \frac{\sigma}{\mu}
\end{equation}

where $\mu$ is the mean and $\sigma$ is the standard deviation (SD) of the transactions’ inter-event times. The burstiness parameter is calculated for all expenditure transactions, fixed and flexible, which reflects how regularly an applicant makes purchases daily. A burstiness value (\emph{B}) of negative one indicates the purchasing pattern is completely stable, zero indicates random behavior and one indicates extreme and unpredicted spikes in expenditure behavior.

Lastly, we are also interested in the role of volatility, particularly regarding applicant’s account balance, income, salary, and expenditure. This is inspired by the growing policy literature highlighting the importance of financial stability, and further intensified by recent phenomena such as ‘zero-hour contracts,’ the gig economy and major economic disruptions like the COVID-19 pandemic. The volatility parameter is computed by measuring the inter-month variation in average account balance (stock), income and salary (flow) and fixed and flexible expenditure amounts. The second-order coefficient variation \citep{Kvalseth2017} is used to avoid issues caused by standard measures of variation which are sensitive to mean and outliers. 

\begin{equation}
    Volatility=\sqrt{\frac{{(\frac{\sigma}{\mu})}^2}{1+{(\frac{\sigma}{\mu})}^2}}
\end{equation}

where $\mu$ is mean and $\sigma$ is SD over the full transaction history. Volatility is expressed as a value between zero and one, with zero indicating low volatility and one indicating high volatility. 

\subsubsection{Financial Distress}

An applicant’s ability to handle financial distress is measured by the following: amount of \emph{debt management} (DM) and \emph{insolvency} expenditure, usage of OD and having RDDs, all of which indicate insufficient funds in the account. The use of Buy Now Pay Later financing (BNPL) is also considered here. BNPL-financing programs (i.e., Klarna, Afterpay, Affirm), also known as point-of-sale loans, have recently become a popular layaway option enabling consumers to buy an item and then split the cost over a few weeks or months with regular installment payments. Usage of such programs may be indicative of situations where an individual does not have sufficient funds in their account to afford paying the full item’s cost at once. However, it may also be the case that due to being typically interest-free credit, it is an attractive option for financially savvy users as well. OD usage is measured by the average number days per month in OD, proportion of months in consistent OD and fees paid. An account is considered in OD when the balance is a negative value for more than one day. RDDs, also known as bounced direct debits, occur when a bank rejects an online check or direct debit, typically when there are insufficient funds in the account to cover the amount requested. This metric is measured by counting the average number of RDD transactions per month. 

\subsubsection{Financial Resilience}

Financial resilience refers to the ability of an individual to withstand unexpected life events that impact his or her income or assets. Financially stressful events can include unemployment, divorce, disability and medical problems. Therefore, to quantify an individual’s financial resilience, features related to account balance value, disposable income and ability to withstand a financial shock of £100 are assessed. To determine an individual’s ‘assets,’ the account balance value is used, with average monthly mean, minimum and maximum recorded. Disposable income which is assumed to be the net amount of money an individual has after paying necessary living expenses, including all fixed expenditures as well as \emph{groceries and housekeeping} expenses (which were originally categorized by the data donor as flexible) is computed as: Income (excl. loan-based income) $-$ all fixed expenditures $-$ \emph{groceries and housekeeping} expenditure. The proportion of months an individual can withstand financial shock is computed as the number of months an individual can withstand financial shock over the total number of months in their account history. 

\subsubsection{Financial Planning}

Financial planning refers to the notion that an individual has a long-term financial strategy in place, represented by the use of insurance, pension, and savings. Holding insurance and consistently paying monthly premiums indicates the ability to protect oneself against unexpected costs and demonstrates financial responsibility. Holding more optional insurance options, such as pet or appliance insurance, may also indicate high financial responsibility. Measures include whether an individual holds any type of insurance and/or pension and the respective monthly average expenditure on each. The monthly average amount put towards savings is also computed based on the categories \emph{debit internal transfers} and \emph{savings \& investments}. 

\subsubsection{Financial Aid}

Financial aid refers to monetary assistance given to certain individuals on a conditional basis such as whether an individual is a carer, has a child, has a disability or is unemployed. As a result, they can receive monthly sums to help alleviate their expenses through carer allowance, child benefits and child tax credits, disability benefits, working tax credit, universal credit and employment support allowance respectively. More information regarding benefit and financial support provided by the UK government can be found on their website (https://www.gov.uk/ browse/benefits). We also include pension income in this category. For each type of benefit, the monthly average value received and its proportion to the individual’s total income are computed. 

\subsubsection{Financial Inclusion}

Financial inclusion refers to the ability of individuals to access useful and affordable financial products and services that meet their financial needs. We consider the use of credit cards and loans as part of this category. However, over-indebtedness is also a key indicator of FV therefore we observe the use of multiple credit or loan providers as evidence of this status. To compute the number of unique credit card providers used, individual providers are identified using NLP techniques, detailed in the Appendix (Figure A.2). Traditional providers refer to ‘high street’ banks such as Capital One, Barclays and HSBC whereas non-traditional providers refer to newer, challenger banks such as Starling and Monzo. This distinction is applied also to loan providers. The monthly average value of credit card payments, loans received, and loans paid are computed.

The aforementioned features are used directly as variables in the prediction and clustering models to generate the results shown in Section \ref{sec:Results}. A descriptive analysis can be found in the Appendix (Figures A.3, A.4) which displays the demographics of our data along target FV indicators and profile features. For example, a monotonic relationship can be seen across variables such as income and expenditure amounts which increase with age, corresponding to the greater number of financial responsibilities and spending that arise. However, later a dip is seen in applicants in the highest age ranges (ages 56 and over), most likely due to unemployment and eventually retirement. 

\section{Empirical Analysis and Results}
\label{sec:Results}

Our aim is to understand to what extent individuals’ FV can be inferred from transaction records, and how these are connected to protected characteristics. To first provide a comprehensive analysis of how one’s financial behavior is associated with individual traits, we use the Pearson correlation coefficient to explore the association between the engineered features with FV indicators and protected characteristics. The complete results are available upon request; however, we briefly highlight select significant correlations (p-value < 0.001). Age is positively correlated to employment length, ability to withstand financial shock and being an owner occupier, likely due to greater stability in one’s employment status thus ensuring more consistent money inflows which help withstand shock, and is negatively correlated with living with parents, likely due to financial freedom. Older individuals have greater total expenditure amounts (\emph{groceries and housekeeping}, \emph{housing} and \emph{utilities} in particular) and higher persistency in categorical spending as well as less burstiness in flexible expenditures. Being female is positively correlated with having child-caring responsibilities and negatively correlated with gambling activities and living with parents, supported by the higher likelihood of having a child. Females tend to spend more in \emph{fashion and beauty} and \emph{groceries and housekeeping} categories along with greater use of BNPL financing options, which may coincide with their growing popularity in the online retail and clothing market \citep{Buynow2021}, as well as higher non-traditional loan usage and number of overall transactions. Being a carer is positively correlated with having a child, likely due to the mediating impact of being female and is also highly correlated with having a disability, which may be due to a carer receiving disability benefits on behalf of the disabled individual for whom they are caring for. Spending on gambling activities is positively correlated with having higher disposable income, implying that those who gamble often gamble with funds remaining after paying for fixed expenses and not necessarily at the risk of going into OD (i.e., spending what they don’t have). Gamblers tend to have higher burstiness in flexible expenditure, indicating that they are more likely to have extreme and unpredicted spikes in expenditure behavior. Overall, we see that the correlations between FV indicators tend to be associated with one another, implying that exhibiting one indicator implies high likelihood of having others. Moreover, the correlations between the FV indicators found in our work reflect the findings of other FV studies \citep{Daud2019, fca2015, oconnor2019}. As a caveat, it is important to note that the identified associations are correlational as opposed to causation driven. Yet these associations are provided to help interpret the predictive models described in the following rather than being prescriptive in their own right. 

\begin{figure}[h]
\includegraphics[width=14cm]{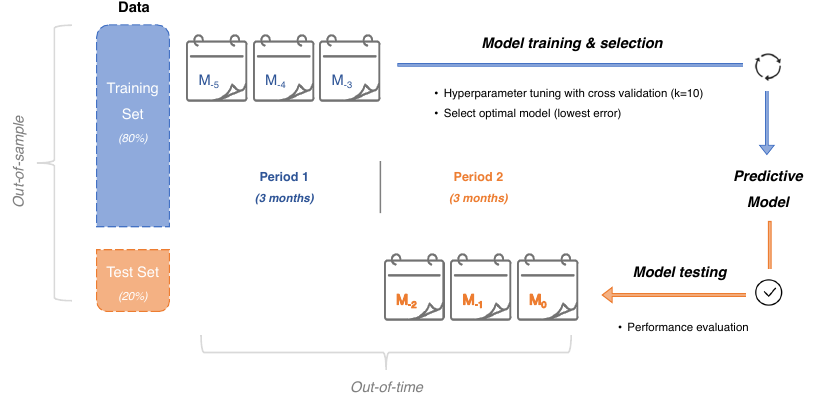}
\centering
\caption{Data and empirical strategy.}
\label{fig:fig4}
\end{figure}

Subsequently, the prediction task is devised as a binary classification problem (Figure \ref{fig:fig4}). In light of exploding data volumes, ML serves as one of the most effective methods in data mining research \citep{Dastile2020}. Therefore, three different classification algorithms are evaluated – Logistic Regression (LR), Random Forest (RF), and Extreme Gradient Boosting (XGBoost). Feature importance is subsequently calculated using the XGBoost model to determine the weightings of the predictive variables. For each model type, the dataset is randomly divided into 80\% training set and 20\% test set while retaining class ratios. Each set spans different three-month time periods thus ensuring out-of-time and out-of-sample testing. The events (e.g., performance along FV indicators) lead to imbalanced datasets where a minority of applicants typically have consistent financial trouble (frequencies are given in the earlier section on FVI). Therefore, to mitigate this imbalance, the majority class is randomly sampled to produce a balanced training dataset and the obtained model is then tested using realistic settings for the testing window. During the training phase, the model parameters are tuned using grid search with ten-fold cross-validation and tested against a 20\% holdout set in the next time window shifted three-months forward. The classifier’s performance is measured as the area under the receiver operating characteristic curve (AUROC) and the feature importances are estimated in terms of (normalized) relative influence. Other performance metrics are also recorded, including accuracy, precision and F1-score. The AUROC metric is focused on given its usefulness in classification scenarios where the trade-off between true positive rate and false positive rate is of vital interest. 

\begin{figure}[h]
\includegraphics[width=14cm]{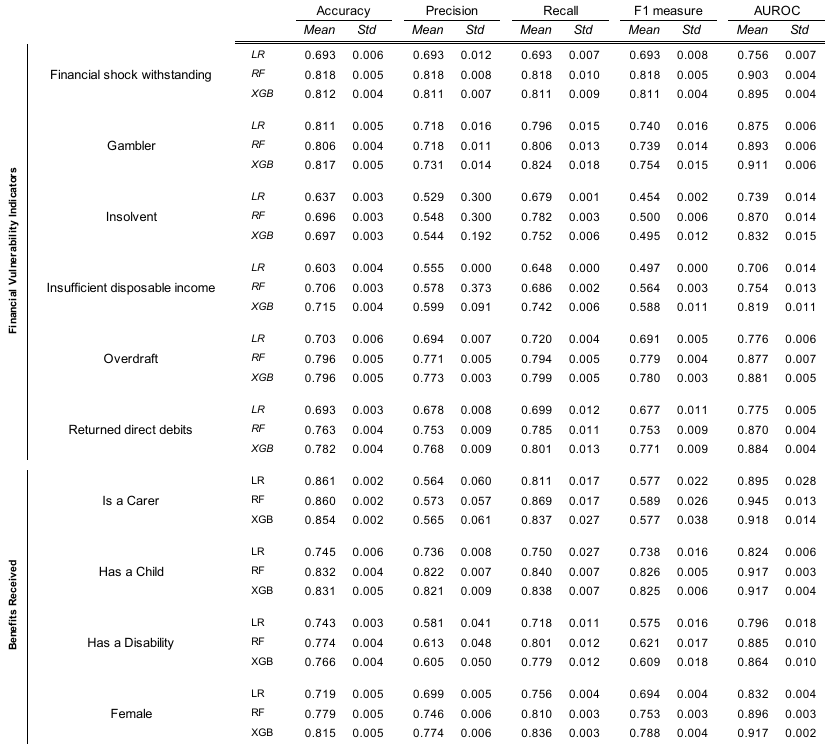}
\centering
\caption{Classification models’ performance. \emph{Note: ML models’ performance (LR = logistic regression, RF = random forest, XGB = XGBoost) evaluated along accuracy, F1 score, precision, recall and AUROC. Real cut-off value is not known and may differ between different lenders therefore AUROC is used for interpretation.}}
\label{fig:fig5}
\end{figure}

\subsection{Classification Performance}

Figure \ref{fig:fig5} displays the predictive performance of LR, RF and XGBoost models for each FV indicator. The performance of the ML models is higher when classifying FV indicators as well as sociodemographic features. The highest performances are obtained with XGBoost to predict whether an applicant is an avid gambler (AUROC = 0.911) and whether they are unable to withstand financial shock (AUROC = 0.895) and the lowest performance when classifying insufficient disposable income (AUROC = 0.819) and insolvency (AUROC = 0.832). This may be due to the significant amount of time needed to reach insolvent status and thus requiring a debt management plan. Reaching this state likely implies the option to OD has already been cancelled by the bank and the individual has applied for payday lenders previously, with high amounts of loan payments or receipts. In other words, reaching insolvency status implies the individual was already financially vulnerable (months or years ago) therefore this indicator is likely not representative of their current financial state (i.e., last six months). Furthermore, predicting whether applicants exhibit protected characteristics is accomplished with decent performance, indicating that “blind” lenders may, in fact, be able to infer personal sociodemographic information even without direct access to it.

\subsection{Feature Importance}

\begin{figure}[h]
\includegraphics[width=10cm]{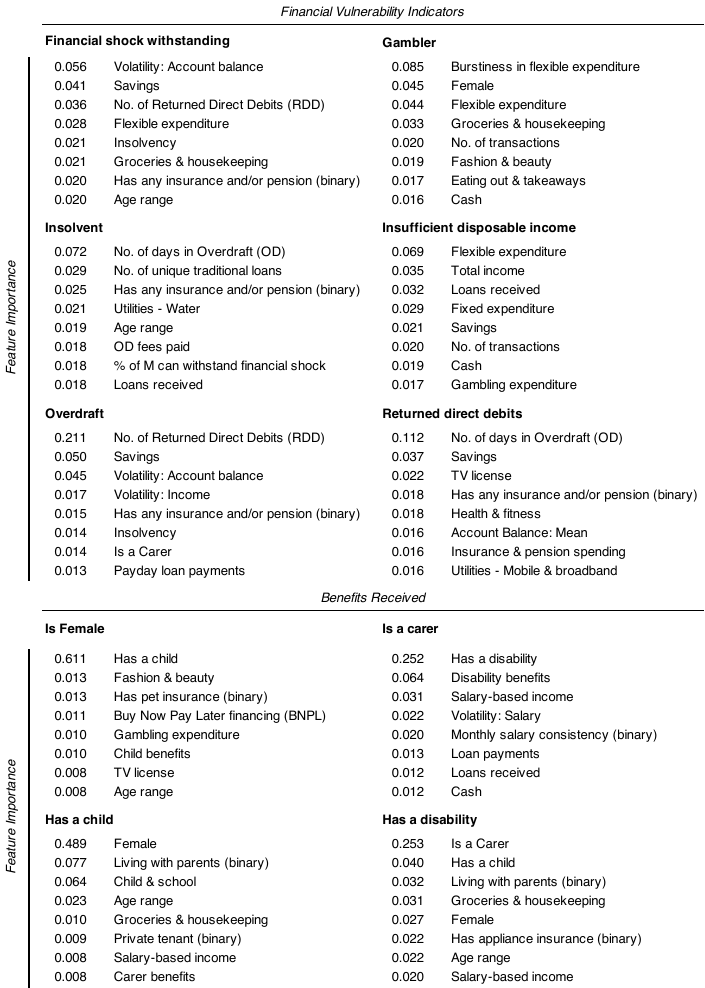}
\centering
\caption{Interpretability of classification models’ performance using feature importance, which is computed as the mean and standard deviation of accumulation of the impurity decrease within each tree.}
\label{fig:fig6}
\end{figure}

To provide interpretability of the outcome, an investigation into the importance of features used by the model is conducted. This enables us to gain insight into which features are most relevant to FV and thereby useful for identifying potential risk of indirect discrimination or proxies. The results are presented in Figure \ref{fig:fig6}, showing the top eight features for each target. We rank all features based on their magnitude of importance using the XGBoost model given their high performance. The importance is computed as the mean and standard deviation of accumulation of the impurity decrease within each tree. 

Supporting the correlation results, the most significant features for predicting FV indicators are typically other indicators, indicating strong linkages between the metrics. For example, predicting whether an individual can withstand financial shock puts importance on features such account balance volatility, savings, number of RDDs and having been insolvent. Predicting insolvency heavily weighs days in OD and number of traditional loans while predicting OD similarly puts importance on RDDs, savings, and volatility in account balance and income. To predict gambling, burstiness in flexible expenditure and gender are important features which is supported by the correlation analysis. Having a child is heavily based on being female, living with parents and \emph{child and school} spending; being a carer on disability benefits, salary and its volatility; and having a disability on carer benefits, likely due to the overlap between those disabled receiving disability benefits directly and those who are carers of disabled individuals receiving benefits on their behalf. Due to the ambiguity of the transaction labels and redacted references, the distinction cannot be further refined. 

\subsection{Clustering Approach}

Clustering is used in our analysis for the task of demonstrating how segmentation can implicitly capture sensitive and protected characteristics, even when they are not included into the analysis. There are multiple applications of clustering in financial data analytics, particularly in behavior analysis \citep{Thompson2020} and marketing, such as Customer Relationship Management (CRM) strategies \citep{Roshan2017} which segment and profile customers according to their needs, desires, or distinguishing characteristics such as age, ethnicity, profession, gender and location or psychographic factors such as shopping behavior, interests, and motivation \citep{Hsieh2004}. With this analysis, we highlight the need for lenders to be concerned not only about the association of certain financial behaviors with a particular sensitive or protected characteristic, but also with the \emph{combination} of several protected characteristics.

\begin{figure}[h]
\includegraphics[width=8cm]{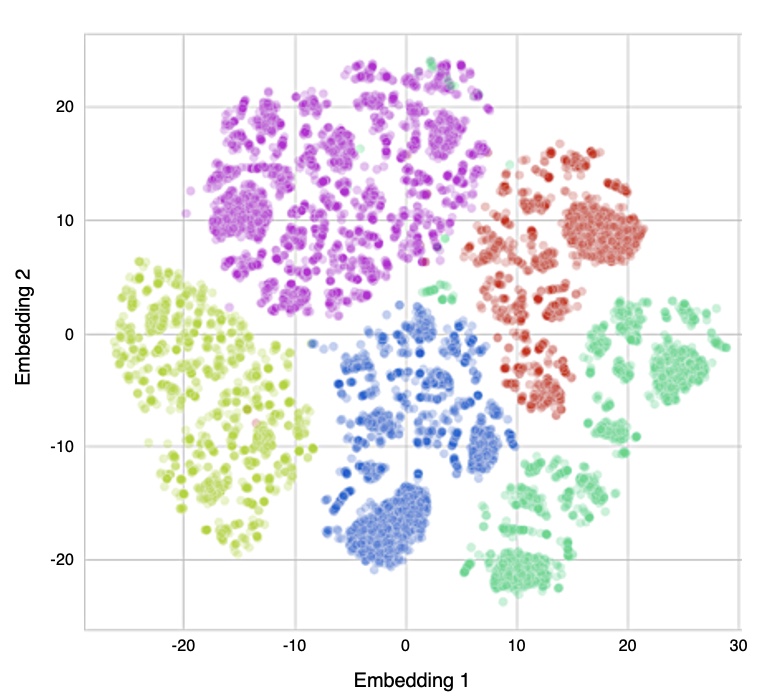}
\centering
\caption{t-SNE visualization for the full dataset by cluster projected onto two embeddings.}
\label{fig:fig7}
\end{figure}

RFM (Recency, Frequency, Monetary) is a widely used method to quantitatively analyze customer behavior based on three dimensions: how recently customers buy, how often they buy, and how much they spend \citep{Cheng2009, Wei2010}. By understanding the history, number and value of customers’ transaction updates, this method can help identify those more likely to respond to certain promotions or for enhancing personalized services. When applied to credit providers, “better” service can include customized loan options, such as loan size, conditions of repayment, or variable interest rates best suited to each applicant’s financial profile. Similarly, we are interested in identifying common profiles of applicants representing different degrees of financial health and financial management ability within our applicant pool. This can be used to determine a range of affordability scenarios for different types of applicants and serve as contextual information for improved affordability assessment measures by lenders. 

\begin{figure}[h]
\includegraphics[width=12cm]{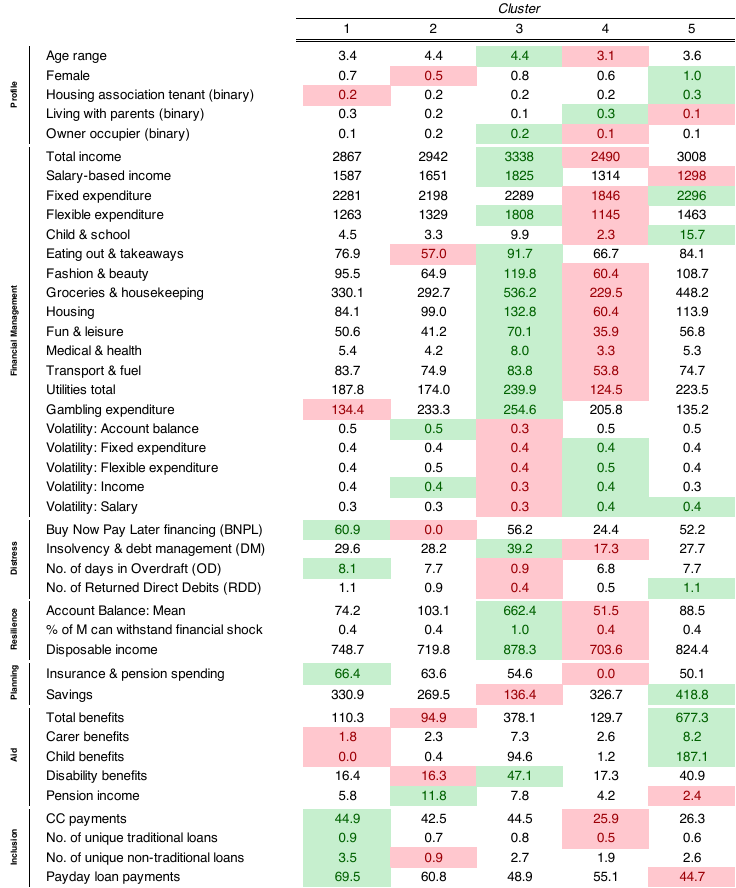}
\centering
\caption{Mean values of the profile and financial management and difficulty features for each cluster, including the features that were not used in obtaining the clusters. \emph{Note: For each feature, lowest values are highlighted in red and highest values are highlighted in green.}}
\label{fig:fig8}
\end{figure}

Principal component analysis is first used to reduce the dimensionality of the data into 15 principal components representing 78.1\% of the cumulative variance. This is followed by the k-means algorithm. To find the optimal number of clusters (k), two assessment measures are applied – the elbow method and silhouette approach – the inflection point and average value maximization respectively point to the use of five clusters. Finally, t-Distributed Stochastic Neighbor Embeddings (t-SNEs) are used for the purpose of visualizing the high-dimensional data. By projecting high dimensional data onto a lower-dimensional space, known as embeddings, t-SNE preserves the local data structure \citep{maaten2008} and forms a non-linear mapping, thus keeping similar data points (i.e., applicants) closer together in the low-dimensional space for visualizing clusters. A perplexity of 300 is used to obtain a stable embedded plot (Figure \ref{fig:fig7}). The data cleaning, feature engineering, correlation, model prediction, clustering algorithm, t-SNE visualization, and analysis are implemented using Python version 3.6. The t-SNE algorithm used for data visualization can be found in the sklearn Python package \citep{Pedregosa2011}. From the two-dimensional embedding map, we can see that there are distinct boundaries between most of the clusters with overlaps between clusters 4 and 5 (green and purple, respectively), however, it is worth noting that higher dimensional embeddings can reveal other higher-order boundaries that may distinguish overlapping clusters. Therefore, the projection from three dimensions to two dimensions, such as in this case for visualization purposes, may create the appearance of overlap. 

\begin{figure}[h]
\includegraphics[width=\linewidth]{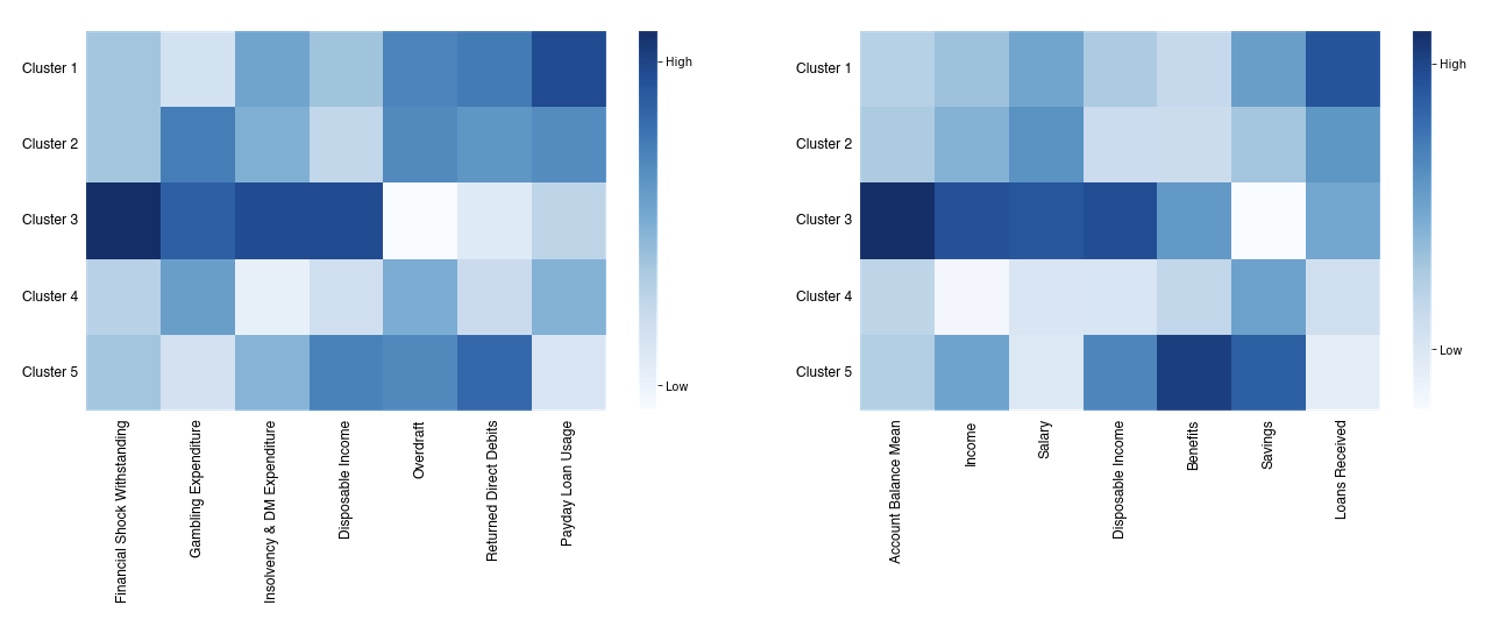}
\centering
\caption{Heatmap display of the key select differentiating features across the five clusters.}
\label{fig:fig9}
\end{figure}

To facilitate the interpretation of the clustering results, the mean values of features of interest (including those that were not used in obtaining clusters, such as demographics) for each cluster are displayed in Figure \ref{fig:fig8}. We first highlight key patterns and defining characteristics to understand the behaviors shared. Clusters 1 and 3 are similar in their credit usage however significantly differ in terms of inflow (income, salary) and outflow (fixed, flexible expenditure). Clusters 2 and 3 are similar in their demographics and income levels, but cluster 3 spends significantly more in expenditures and have lower volatility overall. Cluster 4 is unique with the lowest age and lowest amount of income and salary and most notably, cluster 5 holds a significantly high percentage of female applicants and those with a child. A high-level overview is visualized as a heatmap along key differentiating features, including FV indicators and select inflow and outflow metrics (Figure \ref{fig:fig9}). We note that cluster 3 performs the best along all FV indicators, implying they are the most financially healthy. Next, we summarize in greater detail the unique, distinguishing attributes of each cluster.

\subsection{Applicant Profiles}

While individuals from different groups may appear similar, they are classified based on subtle differentiating factors determined by the clustering algorithm. Therefore, we narrate “profiles” of applicants to ease discussions and better understand the clusters as real people, not simply data points. \emph{Cluster 1 – The Credit User} (19.4\% of applicants): These applicants use significant amounts of financial aid and credit options in the form of loans, the use of a payday lender, credit cards and BNPL financing. Relative to other clusters, this group has the highest credit card usage and payments, across both traditional and non-traditional providers in addition to the highest loan usage, both traditional and non-traditional, particularly with a payday lender. The amount of loans paid makes up 16\% of their income, the highest proportion relative to other clusters. These applicants have a high volatility in average account balance, highest proportion of days in OD and highest number of RDDs. However, they are less likely to have dependents with minimal disability or carer benefits as well, and are more likely to live with their parents, thereby reducing their housing costs.

\emph{Cluster 2 – The Financially Resilient} (15.9\% of applicants): These applicants represent the second oldest age group with the highest insurance and pension spending but with low to no benefits received. They have a low likelihood of having a child, however, receive the second highest amount of pension compared to cluster 3, coinciding with their average age range. This group has low average disposable income along with low amounts of voluntary expenditures in categories such as \emph{eating out and takeaway} as well as \emph{fashion and beauty}. They also hold a low likelihood of non-traditional loan usage, however, have high usage of a payday lender. 

\emph{Cluster 3 – The Financially Secure} (19.0\% of applicants): These applicants represent the most financially secure and healthy group of individuals with the highest inflows and highest outflows. They represent the oldest age group which correlates with their higher employment length and may explain the financial security evidenced by having the highest average account balance and disposable income relative to other clusters. They are also able to withstand financial shock nearly 100\% of the time. This group is characterized by a consistent monthly salary, lowest number of days in OD and lowest volatility in account balance, income, and salary. They also receive a high proportion of disability benefits and pension-based income. Alongside higher inflows, this group also displays high expenditure amounts, particularly with the highest flexible expenditure amounts across all spending categories (e.g., \emph{eating out, fashion, fun and leisure, subscriptions, charity, housing, medical and health, groceries, gambling} and \emph{utilities}). This group holds the largest proportion of owner occupiers which coincides with their financial security.
 
\emph{Cluster 4 – The Young and Challenged} (16.7\% of applicants): These applicants represent the youngest group with high volatility in salary and income but correspondingly with the lowest expenditures, both fixed and flexible. This cluster has the lowest earnings relative to other groups but also the lowest spending with high burstiness in flexible spending, indicating uncertainty and a more volatile lifestyle. They are unable to withstand financial shock for most parts of the year (65\% of months) along with their low average account balance of £51. Their salary represents a smaller proportion of their overall income with higher likelihood of weekly salary inflows (as opposed to the monthly standard), implying alternative income sources. This group receives minimal benefits, most likely due to their young age, with the lowest usage of traditional loans but higher usage of non-traditional loans, which may reflect younger adults shifting towards FinTech products. This group has the highest likelihood of living with their parents and lowest of being an owner occupier, which accounts for their significantly low \emph{utilities} expenditure, particularly in energy usage. 

\emph{Cluster 5 – The Beneficiary} (29.1\% of applicants): These applicants hold the highest proportion of females relative to other clusters alongside having the highest likelihood of holding responsibilities such as having a child or being a carer. As a result, this group subsidizes their living costs with a significant proportion of benefits, making up 28\% of their overall income. Their categorical spending skews largely towards \emph{child and school} expenditure while saving \emph{housing} costs by being a housing association and/or council tenant. They have minimal \emph{gambling} expenditure, likely due to the higher proportion of females. This group is also characterized with the highest number of RDDs with a high proportion of days spend in OD and high BNPL usage. However, they are not as likely to use loans, a payday lender or receive pensions. Interestingly, they are characterized with the highest volatility in salary but lowest volatility in income relatively, implying significant non-salary income support, mainly in the form of benefits. Therefore, when salaries are impacted for this demographic, they are still able to accrue consistent income from other public transfers. 

\begin{figure}[h]
\includegraphics[width=8cm]{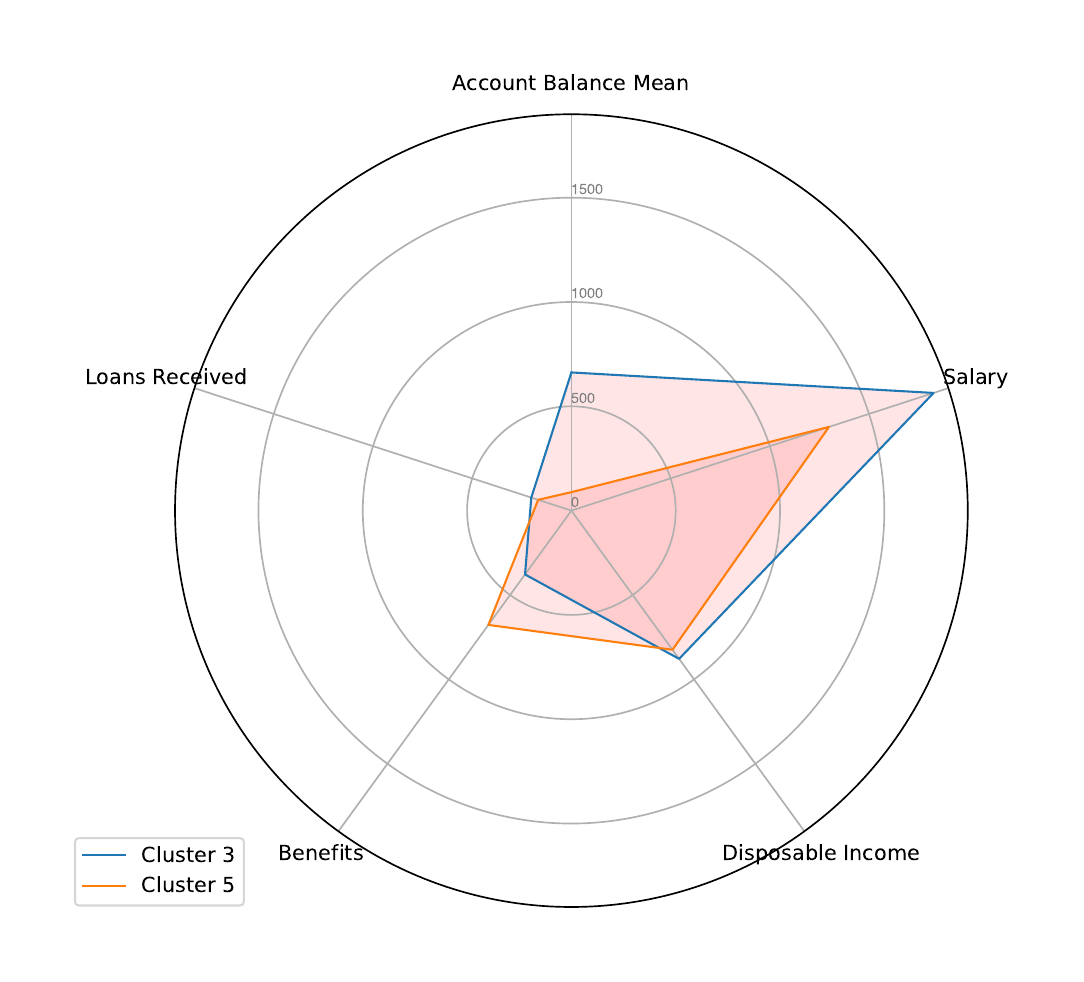}
\centering
\caption{Radar chat comparison between cluster 3 \emph{(The Financially Secure)} and cluster 5 \emph{(The Beneficiary)}.}
\label{fig:fig10}
\end{figure}

Let us consider in greater detail clusters 3 and 5 (Figure \ref{fig:fig10}), comparing two distinct groups in terms of socio-demographic profile and financial behavior. We can see that while salary inflows differ significantly, both clusters have comparable disposable income. This can be attributed by the significant proportion of benefits received by cluster 5 which compensates for their lack of salary, thus highlighting the importance of child and unemployment benefits to their financial wellbeing. With benefits making up 28\% of their income on average, these individuals would be unable to cope with necessary life expenses without such support or if they are affected by the loss of supplementary income. This poses the question of whether lenders should be taking benefits and proportion of these benefits to overall income into consideration within their affordability criteria when assessing applicants. And if not, whether this may adversely affect women with children and thus risk indirectly discriminating against this group of individuals. 

\section{Discussion}
\label{sec:Discussion}

Our study has involved a variety of approaches analyzing an Open Banking dataset to explore financial behaviors and their implications on FV and fairness. We propose a methodology for identifying and assessing FV using transaction data, including six major indicators for differentiating levels of financial stability, capacity and management ability. The engineered features can be used to improve the performance of predictive models regarding individualized credit risk, with better insight into FV warning signs particularly for at-risk populations. This may be useful for alternative lenders looking to expand their customer base into underbanked populations. The results suggest that transaction data can be highly predictive of future FV, three months ahead of model application. The kind of analysis described here can be used to judge new loan applicants for creditworthiness, where the given models are applicable to both applicants who can provide traditional credit data enhanced with transaction history (hybrid model) as well as applicants with limited or no credit scoring history (transaction-only model). 

We’ve shown that protected characteristics can also be inferred, even when removed from the underlying data. This renders the fairness via mandated “blindness” approach (also known as “fairness through unawareness” \citep{Dwork2011}, which naively ignores all protected attributes, as futile in the current data environment. We would like to point out the risk of certain behavioral attributes serving as proxies for these attributes, enabling the prediction of omitted attributes through other permitted features. Lastly, the clustering results reveal two major implications. Firstly, they illustrate that decision-makers should remain concerned about the combination of protected characteristics with certain financial behaviors. Secondly, these results can help policymakers and practitioners form a profile of the type of consumers who are most at risk of “sliding” from a state of low to high FV over time and in need of the most support. For example, signaling one’s financial health can be done incrementally before payment deadlines to inform the user or lending institution if preemptive remedies are needed before missing payments or defaulting. Doing so also helps address recent calls urging stakeholder groups to pay greater attention to the characteristics of consumers “trending toward vulnerability” \citep[p. 427]{oconnor2019}. Recent household surveys reveal that one in five households have depleted their savings, fallen behind on housing payments, or are experiencing difficulty paying their debts, buying groceries, and paying utilities as a result of the COVID-19 pandemic \citep{centeronbudget2020}, stressing the timeliness and gravity of this issue. In the future, we would like to explore how decision making can be personalized for different applicant clusters. This has important implications for individuals as opportunities for data sharing in return for opportunities (i.e., a loan) grow, and organizations look to draw more detailed insights from their customers for marketing, retention or risk management purposes. 

Some unique highlights are also mentioned. Firstly, our data set encompasses a particular demographic in the UK, who majorly work in the public sector and tend to be more financially challenged than average. This is not only evidenced in the data (i.e., salary and income amounts) but via selection bias of the applicants received. The dataset was collected from applicants who voluntarily applied to an alternative lender which differentiates itself by not requiring a standard credit check, thus attracting a niche demographic. While this may be seen as a limitation, we view it as an opportunity to shed light on individuals often lost at tail ends of the general population; or those holding a different distribution over select features and thus a different relationship with the predicted label relative to others. Typically, models fitted to the majority which look to minimize overall error can result in representation bias – a key cause of unfairness \citep{Mehrabi2019}. Secondly, there are limited studies that utilize standalone Open Banking data. Researchers have shown improved predictive capability by supplementing traditional credit models with alternative data \citep{Djeundje2021, Gambacorta2019, oskardottir2019}, however, there are no studies in the context of FV with consideration of fairness, to the best of our knowledge. This is also the first that provides insight tailored around an at-risk demographic. However, we believe our analytical methods can be applied to other populations with transaction-based datasets and hope our use of routine and temporal patterns for characterizing financial behavior will inspire future research. 

We note that this study contains some limitations that could be addressed with future research. First, we do not establish causality, rather the findings demonstrate associations that may reflect causality or comorbidity – both of which are of concern. Causality would indicate, for example, higher levels of gambling increase one’s risk of FV. Comorbidity, however, would indicate that individuals who are susceptible to such negative outcomes due to alternative factors are more likely to be drawn to gambling; for example, where having a bounced check leads to gambling as a means to pay off debt. In reality, the observed effects are likely a blend of both effects. Furthermore, while OD is considered a FV indicator in this study, it is worth noting the possibility of individuals without OD may, in fact, be more financially vulnerable because they do not have a facility to use in the case of financial shocks or alternatively do not have the option of OD to begin with because their bank considers them high risk. Further work is needed to measure the extent to which FV is driven by causal mechanisms. Nonetheless, this longitudinal transaction-based approach informs the current FV debate. Vulnerability characterization using financial behavioral markers, covering consumption, savings, and monetary inflow can be insightful for economic research and public policy to help identify individuals who may be more sensitive to income shocks; for example, distinguishing those paid on a consistent monthly basis versus those self-employed or largely living off benefits. It also informs how much households would have to reduce flexible expenditures or savings to maintain basic, necessary consumption levels. In that sense, this work adds to the existing literature on the use of Open Banking data as a means for assessing and possibly preventing FV.

\subsection{Future Work and Applications}

Our work ultimately looks to help promote financial sustainability. Naturally, this study is followed by asking what level of shock each consumer type can withstand and formulating strategies to increase the financial prospects of select groups. Future work looks to explore stress testing methods to discern how robust users are to certain types and levels of financial shock – a critical concern of credit providers but with applications to other financial institutions such as mortgage and insurance providers \citep{Marron2007, Mester1997}. For example, a stable, responsible individual may be able to overcome a financial shock successfully with little, temporary help as opposed to a shopaholic or avid gambler who may be tempted to spend funds irresponsibly into OD. Understanding what form of credit would be beneficial for an individual, at what point in time, would be informative to lenders. Due to the recency of the applications, we are unable to discern repayment behavior therefore we plan to examine the linkages between FV indicators and repayment habits, while looking for evidence on whether we can nudge any of the noted behaviors. 

In real-world settings, this information can be used to determine whether the demand for credit and capacity of those in need match the supply of options available in the current financing market. If it is not possible for this demographic to sustain the market rates, this begs the question of whether regulatory agencies should act to ensure more affordable lenders are available or publicly sponsored. Furthermore, with the appropriate checks and balances established, the observations presented here could be used to provide individual feedback and nudge healthier decision making. For example, financial support can be released in flexible ways, such as frequent smaller payments rather than a lump sum with real-time feedback provided via transactional data. If a consumer acts reasonably, the lender can decide to release additional funds or relax interest rates, enabling disadvantaged individuals to rise out of their debt cycle. This can act as a “financial rehab” of sorts for those deemed higher risk along traditional credit metrics but desire an opportunity to improve their financial standing over time, rather than remain limited to unaffordable or fringe lenders. 

While all the data used in this study is anonymized and at no point are individuals identifiable in its undertaking, we remain highly cognizant of the implications surrounding behavior-to-outcome associations. As the data landscape grows more complex, data points are no longer singular but can have multiple contextual implications, fueling concerns over fairness, discrimination as well as privacy. We highlight that certain groups of individuals may be unknowingly or unintentionally disadvantaged even when seemingly ‘neutral’ data is utilized. In most cases, individuals will not expect or even be aware of the fact that their consumption data may reveal personal factors. Realistically even the most motivated users will find it increasingly difficult to accrue the knowledge required to make self-interested decisions which trade-off against immediate and tangible benefits that may come with data sharing. Therefore, we conclude by challenging the claim that all algorithms are neutral and caution against an “all data is credit data” approach \citep{Aitken2017}. As the potential for discriminatory harms magnify in proportion to the technological advancements, we provide a word of caution to lenders using Open Banking data where the risk of abuse can easily outweigh its potential benefits without proper oversight.

\paragraph{Acknowledgements}
The authors are grateful to the data provider who has chosen to remain anonymous. S. Kim is grateful to Baillie Gifford for funding. 

\graphicspath{ {./images/} }

\newpage
\bibliographystyle{unsrtnat}
\bibliography{references}  

\newpage
\appendix{\Large \textbf{Appendix}}
\label{sec:Appendix}
\setcounter{figure}{0}
\renewcommand\thefigure{A.\arabic{figure}}

\begin{figure}[h]
    \includegraphics[width=5cm]{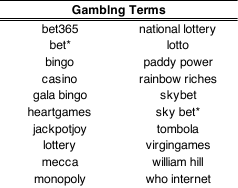}
    \centering
    \caption{Gambling-related terms used for NLP processing of transaction references. \emph{Note: If outgoing transaction references contained any of these terms in its text, they are considered gambling-related.}}
    \label{fig:figa1}
\end{figure}

\begin{figure}[h]
    \includegraphics[width=14cm]{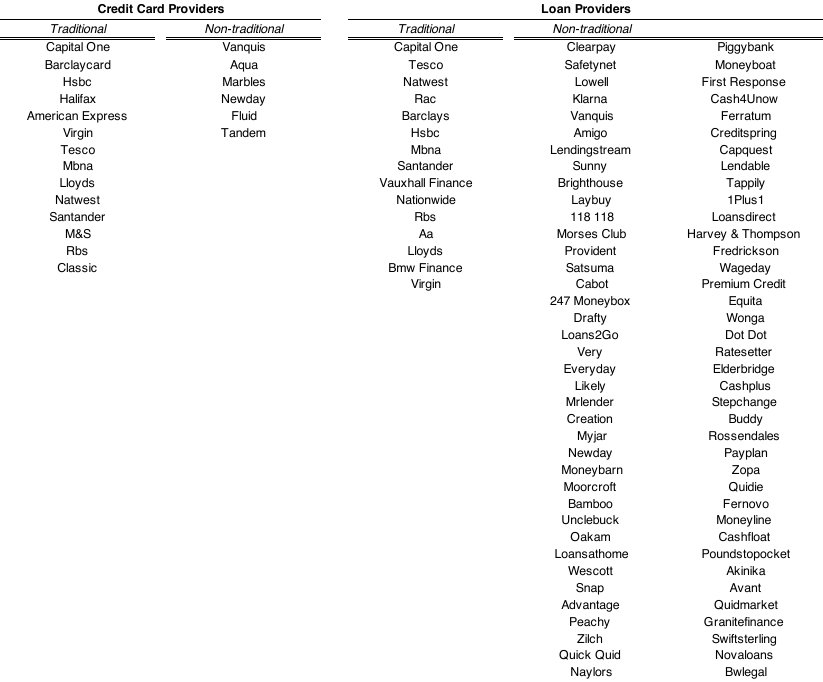}
    \centering
    \caption{Selection of credit card and loan providers based on traditional (high-street banks) or non-traditional (newer banks). \emph{Note: If incoming (credit) or outgoing (repayment) transaction references contained these terms in its text, they are correspondingly categorized.}}
    \label{fig:figa2}
\end{figure}

\begin{figure}[h]
    \includegraphics[width=14cm]{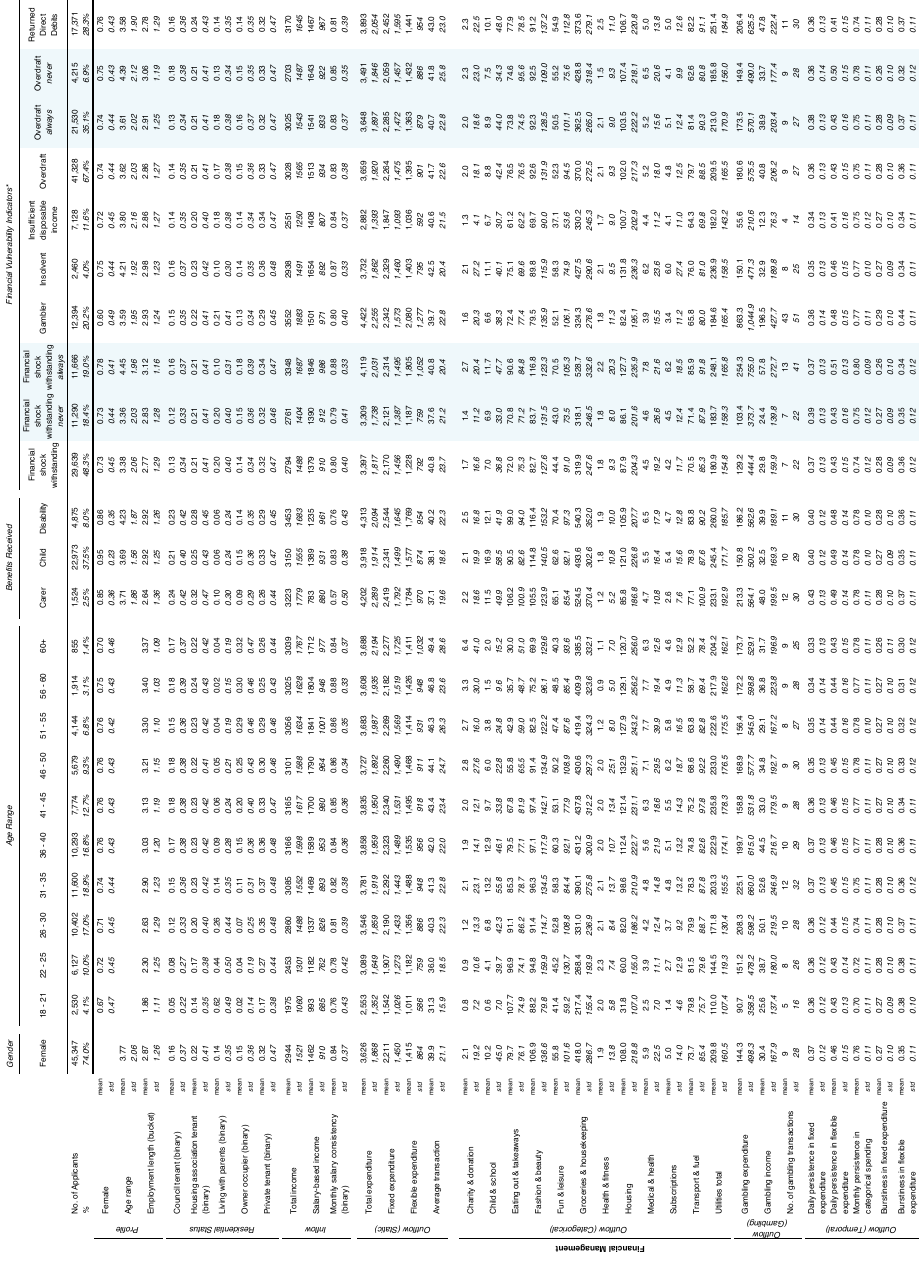}
    \centering
    \caption{Summary statistics of applicants’ pool (Part 1). \emph{Note: Table reports the mean and SD (in parentheses) of the variables. Profile-related features are provided as inputs by the applicants themselves during the application process. All other features are constructed based on the transaction data collected at the time of application via a third-party Open Banking API provider. All attributes represent monthly averages with amount values in £/GBP unless noted otherwise.}}
    \label{fig:figa3_1}
\end{figure}

\begin{figure}[h]
    \includegraphics[width=14cm]{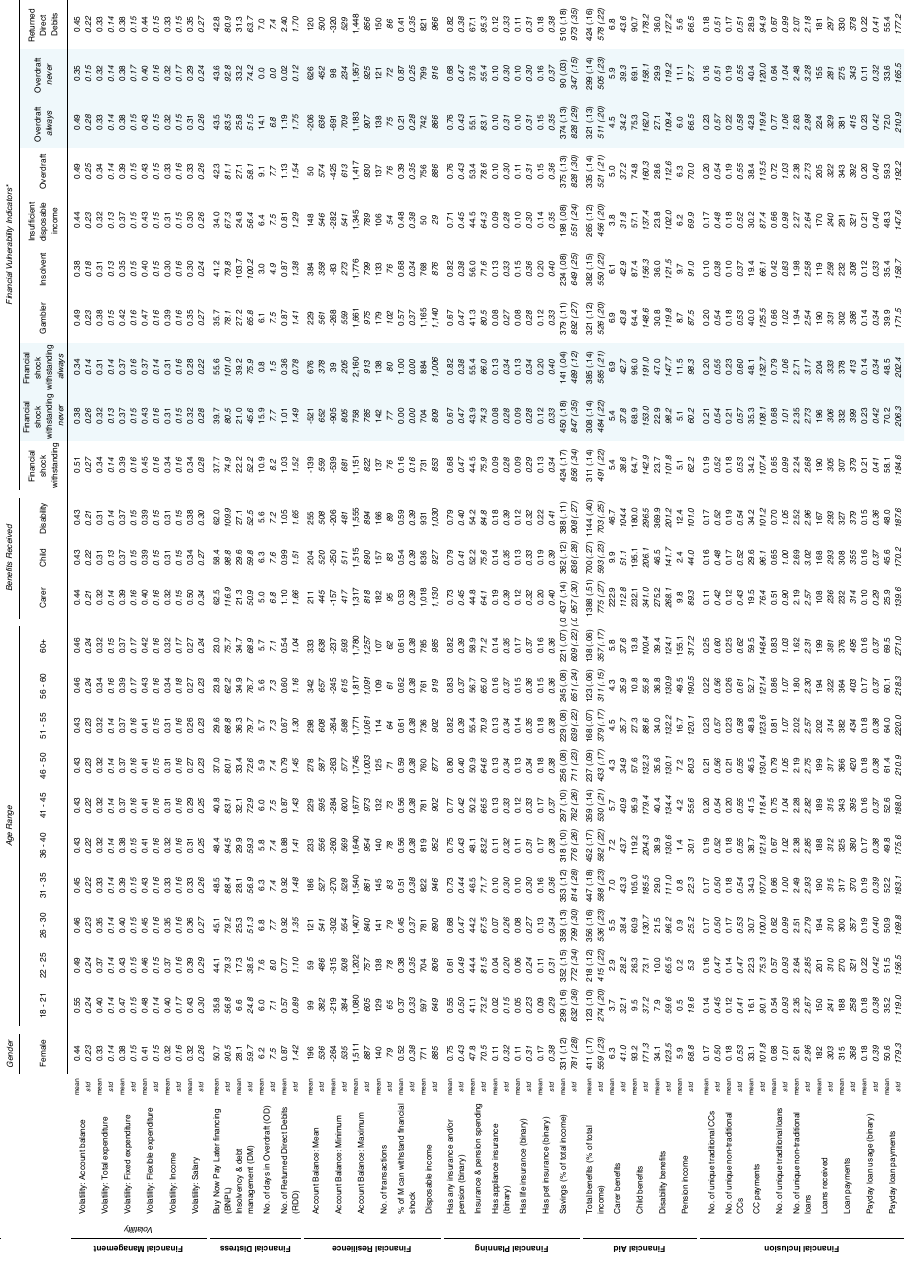}
    \centering
    \caption{Summary statistics of applicants’ pool (Part 2). \emph{Note: Table reports the mean and SD (in parentheses) of the variables. Profile-related features are provided as inputs by the applicants themselves during the application process. All other features are constructed based on the transaction data collected at the time of application via a third-party Open Banking API provider. All attributes represent monthly averages with amount values in £/GBP unless noted otherwise.}}
    \label{fig:figa3_2}
\end{figure}

\end{document}